\documentclass[11pt]{article}
\usepackage[a4paper,margin=1in]{geometry}
\usepackage[T1]{fontenc}
\usepackage{amsmath,amssymb,amsthm}
\usepackage{graphicx,booktabs}
\usepackage[hidelinks]{hyperref}
\usepackage[capitalize,noabbrev]{cleveref}

\newtheorem{theorem}{Theorem}[section]
\newtheorem{proposition}[theorem]{Proposition}
\newtheorem{lemma}[theorem]{Lemma}
\newtheorem{corollary}[theorem]{Corollary}
\theoremstyle{definition}
\newtheorem{definition}[theorem]{Definition}
\newtheorem{remark}[theorem]{Remark}

\newcommand{\R}{\mathbb R}
\newcommand{\T}{\mathsf T}

\DeclareMathOperator{\lse}{lse}

\newcommand{\Linf}{\mathcal L_\infty}

\title{Conic Structure of the Cross-Entropy Loss at the Final Layer and Its Absence in
Rectifier Backbones}
\author{\"O. N. Gerek\thanks{Department of Electrical and Electronics Engineering,
Eski\c{s}ehir Technical University, Eski\c{s}ehir, Turkey.}}
\date{}

\begin{document}
\maketitle

\begin{abstract}
Slices of the loss along lines and planes in parameter space are widely used to visualize
deep networks. They often show a single basin with nearly straight walls, which raises the
question of whether such a loss is a cone rounded at its apex, or a lower envelope of
several cones that meet along creases. We answer this question for softmax cross-entropy
at two levels of a classifier. Over the weights of an affine final layer, the loss equals a
polyhedral cone plus an offset that decreases monotonically and exponentially along every
ray, towards a limit in closed form. The distance at which the asymptotic wall reaches the
height of the starting point, which we call the core radius, equals the mean surprisal of
the classes that the direction favours, divided by the slope of the wall. Over the
parameters of a rectifier backbone, the loss along a line has a kink wherever the
activation pattern of a sample changes. Using the dimension theory of semialgebraic sets,
we show that, along almost every line, the samples that switch at one point share an
irreducible factor of their pre-activation polynomials. Under a coprimality condition,
each kink therefore carries the share of one sample among $n$, so creases comparable to
the slope do not form. Measurements on ResNet-18 and ResNet-56 trained on CIFAR-10
illustrate both results, with a test that bounds the size of any crease that sampled loss
values can hide.
\end{abstract}


\section{Introduction}
\label{sec:intro}

The loss of a deep network is a function of millions of parameters, and it is commonly
studied along lines and planes through a trained point
\cite{goodfellow2015qualitative,li2018landscape,garipov2018mode,draxler2018barriers}. Such
slices often show a single basin whose walls become nearly straight away from the minimum,
and whose surface carries small wrinkles. Two pictures suggest themselves. In the first,
the loss around a trained point is a cone rounded at its apex, so that it grows linearly
along every ray, with a slope that depends on the direction. In the second, a wrinkled
loss surface is a lower envelope of several cones, which meet along creases where the
slope jumps by an amount comparable to the slope itself. Both pictures are statements about
the shape of a function, and neither has been examined exactly.

The present paper examines them for softmax cross-entropy at two levels of a classifier.
At the final layer, the logits depend linearly on the weights, and the first picture holds
in an exact sense. In the backbone, the parameters reach the logits through rectifications,
and we prove that the second picture does not hold there, because along almost every line
the kinks of the loss are too small and too numerous to form creases.

\subsection{Main results}
\label{sec:results}

For an affine final layer, \cref{prop:finallayer} places softmax cross-entropy over the
weights within a band of constant width around a polyhedral cone, whose scaled level sets
converge to one convex set. The cone is the recession function of the loss. \Cref{thm:ray}
sharpens the band into an exact expansion along every ray, in which the offset from the
cone decreases monotonically and exponentially to a limit in closed form. \Cref{cor:core}
then gives the core radius of the rounded cone as the mean surprisal of the classes that
the direction favours, divided by the slope of the wall. The core radius is small at the
origin with equal biases, where the softmax is uniform, and large at a trained point, where
the softmax is confident.

For a rectifier backbone, the loss along a line in parameter space has a kink wherever
the activation pattern of a sample changes (\cref{prop:kinks}). \Cref{thm:generic} uses the
dimension theory of semialgebraic sets to show that, for almost every direction, the
samples that switch at one point share an irreducible factor of their pre-activation
polynomials. Under a coprimality condition, each kink is therefore the share of a single
sample, weighted by $1/n$. A crease of an envelope of cones is a kink comparable to the
slope itself, and forming it would require a large fraction of the samples to switch at
the same point. \Cref{lem:floor} bounds the size of any crease that sampled loss values can
hide, and \cref{rem:walk} explains the roughness that many small kinks produce.

\Cref{sec:numerics} illustrates each statement on residual networks trained on CIFAR-10
\cite{krizhevsky2009cifar}. On slices through the final layer of ResNet-18, the exact
expansion reproduces the wall slopes and core radii, and it shows that a rounded cone
fitted to sampled values overestimates the core radius. On lines and planes through the
whole backbone of ResNet-56, with and without skip connections, each slice has a single
basin, and the bound of \cref{lem:floor} excludes creases comparable to the slope on every
line.

\subsection{Related work and notation}
\label{sec:related}

Three lines of work meet these results. The first concerns the implicit bias of gradient
descent on separable data, whose iterates move out along directions that classify every
sample correctly \cite{soudry2018implicit,ji2019implicit}. These works follow the
trajectory of an algorithm. The present results describe the loss itself, along every ray
from any starting point, including the directions that do not separate the data. The
second line is convex analysis, in which the recession function describes the growth of a
convex function along a direction \cite{rockafellar1970convex,auslender2003asymptotic}.
The recession function of the log-sum-exp is classical, and \cref{thm:ray} adds the exact
offset along every ray, its monotone and exponential convergence, and the closed form of
the core radius. The third line counts the linear pieces of rectifier networks as
functions of their input \cite{montufar2014regions,hanin2019regions} and visualizes loss
surfaces along lines and planes
\cite{goodfellow2015qualitative,li2018landscape,garipov2018mode,draxler2018barriers}.
\Cref{thm:generic} concerns the loss as a function of the parameters and explains why such
slices show many small kinks and no creases. A companion paper studies a conic
classification head in feature space, where the conic form is present by construction
\cite{gerek2026margins}.

Throughout, $\|\cdot\|$ denotes the Euclidean norm, $[K]=\{1,\dots,K\}$ is the set of
classes, and a training set has $n$ samples $x_i$ with labels $y_i\in[K]$. \Cref{sec:final}
treats the final layer and \cref{sec:backbone} the backbone. \Cref{sec:numerics} reports
the measurements and \cref{sec:discussion} discusses them. The supplementary material
contains numerical examples of the theorems on small instances, the training protocol and
further measurements.

\section{The final layer}
\label{sec:final}

This section treats the loss as a function of the weights of an affine final layer, with
the backbone, and hence the features, held fixed. The logits then depend linearly on the
weights, which makes exact statements possible. Let the
features $z_1,\dots,z_n\in\R^d$ with labels $y_1,\dots,y_n$ be fixed, and let the final layer
produce the logits $\langle W_k,z_i\rangle+b_k$, where $W_k$ is the $k$-th row of
$W\in\R^{K\times d}$. The empirical risk is
\begin{equation}
\mathcal L(W)=\frac1n\sum_{i=1}^n\Bigl[\lse\bigl(Wz_i+b\bigr)-\langle W_{y_i},z_i\rangle-b_{y_i}\Bigr],
\qquad \lse(x)=\log\sum_{k=1}^Ke^{x_k}.
\label{eq:LW}
\end{equation}
Its recession function, which describes how $\mathcal L$ grows far from any given point
\cite{rockafellar1970convex,auslender2003asymptotic}, replaces the log-sum-exp by the
maximum and discards the bias,
\begin{equation}
\Linf(D)=\frac1n\sum_{i=1}^n\Bigl[\max_k\langle D_k,z_i\rangle-\langle D_{y_i},z_i\rangle\Bigr],
\qquad D\in\R^{K\times d}.
\label{eq:recession}
\end{equation}
We call $\mathcal F=\{D:\Linf(D)=0\}$ the floor and $S_1=\{D:\Linf(D)\le1\}$ the unit
level set of the cone.

\subsection{A band around a polyhedral cone}
\label{sec:band}

The first result compares the loss with its recession function over the whole weight
space.

\begin{proposition}[The final-layer loss is a rounded polyhedral cone]
\label{prop:finallayer}
Let $W_0$ be any weight matrix, let $s^0_{ik}=\langle W_{0,k},z_i\rangle+b_k$ be its
logits, and let
\begin{equation*}
m_-=\frac1n\sum_{i=1}^n\Bigl(\min_ks^0_{ik}-s^0_{iy_i}\Bigr),\qquad
m_+=\frac1n\sum_{i=1}^n\Bigl(\max_ks^0_{ik}-s^0_{iy_i}\Bigr)+\log K .
\end{equation*}
\begin{enumerate}
\item[(i)] The function $\mathcal L$ is convex. The function $\Linf$ is convex,
positively homogeneous of degree one, piecewise linear and non-negative, and the floor
$\mathcal F=\{D:\langle D_{y_i}-D_k,z_i\rangle\ge0\text{ for all }i,k\}$ is a polyhedral
convex cone.
\item[(ii)] For every $W$,
$\Linf(W-W_0)+m_-\le\mathcal L(W)\le\Linf(W-W_0)+m_+$.
\item[(iii)] For every direction $D$, the chord slope
$\bigl(\mathcal L(W_0+tD)-\mathcal L(W_0)\bigr)/t$ is nondecreasing in $t>0$ and tends to
$\Linf(D)$.
\item[(iv)] For every level $c>m_+$,
\begin{equation}
\Bigl(1-\frac{m_+}{c}\Bigr)S_1\;\subseteq\;\frac1c\bigl(\{W:\mathcal L(W)\le c\}-W_0\bigr)
\;\subseteq\;\Bigl(1-\frac{m_-}{c}\Bigr)S_1 ,
\label{eq:levels}
\end{equation}
so the level sets of $\mathcal L$, scaled by $1/c$, converge to $S_1$ as $c\to\infty$.
\end{enumerate}
\end{proposition}

\begin{proof}
(i) The log-sum-exp is convex and each logit is linear in $W$. Each summand of $\Linf$ is a
maximum of linear functions of $D$ minus a linear function, and it is non-negative because
the maximum is at least the term of the true class. (ii) With $W=W_0+D$ and $q_i=Dz_i$, the
$i$-th summand of $\mathcal L(W)$ is $\lse(s^0_i+q_i)-s^0_{iy_i}-q_{iy_i}$. The bounds
$\max_k x_k\le\lse(x)\le\max_kx_k+\log K$ and $\max_k q_{ik}+\min_k
s^0_{ik}\le\max_k(s^0_{ik}+q_{ik})\le\max_k q_{ik}+\max_k s^0_{ik}$ place the summand within
$[\min_ks^0_{ik}-s^0_{iy_i},\,\max_ks^0_{ik}-s^0_{iy_i}+\log K]$ of $\max_kq_{ik}-q_{iy_i}$.
(iii) The chord slope of a convex function along a ray is nondecreasing, and by (ii) it
differs from $\Linf(D)$ by at most $(m_+-m_-)/t$. (iv) Since $c>m_+>m_-$, positive
homogeneity gives $\{D:\Linf(D)\le c-m_\pm\}=(c-m_\pm)S_1$, and the two inequalities of (ii)
give the two inclusions.
\end{proof}

The loss of the final layer thus stays within a band of constant width $m_+-m_-$ around a
polyhedral cone with its apex at any chosen $W_0$. At $W_0=0$ the width is the spread of
the biases plus $\log K$, and at a trained point, where the logits are large, the band is
much wider. The cone $\Linf$ belongs to the same class of convex, positively homogeneous and
piecewise linear functions as the $\ell_1$ term of the polyhedral conic functions
\cite{gasimov2006pcf}, so the conic form of the final layer is polyhedral.

\subsection{Exact expansion along a ray}
\label{sec:ray}

Along a single ray, the loss has an exact expansion. For a direction $D$, write
$q_{ik}=\langle D_k,z_i\rangle$, $\bar q_i=\max_kq_{ik}$, and let
$A_i=\{k:q_{ik}=\bar q_i\}$ be the set of classes that $D$ favours on the sample $i$. When
$A_i\ne[K]$, let $\omega_i=\bar q_i-\max_{k\notin A_i}q_{ik}>0$, and let
$p^0_i(A)=\sum_{k\in A}e^{s^0_{ik}}/\sum_{k}e^{s^0_{ik}}$ be the probability that the
softmax at $W_0$ assigns to a set $A$ of classes on the sample $i$.

\begin{theorem}[Exact expansion along a ray]
\label{thm:ray}
For every weight matrix $W_0$, every direction $D$ and every $t\ge0$,
\begin{equation}
\mathcal L(W_0+tD)=t\,\Linf(D)+b(D)+R_D(t),\qquad
b(D)=\frac1n\sum_{i=1}^n\Bigl[\log\sum_{k\in A_i}e^{s^0_{ik}}-s^0_{iy_i}\Bigr],
\label{eq:ray}
\end{equation}
where the remainder $R_D$ is convex and nonincreasing, $R_D(0)=-\frac1n\sum_i\log
p^0_i(A_i)$, and
\begin{equation}
0\le R_D(t)\le\frac1n\sum_{i:\,A_i\ne[K]}\bigl(K-|A_i|\bigr)
\exp\Bigl(\max_{k\notin A_i}s^0_{ik}-\max_{k\in A_i}s^0_{ik}-t\,\omega_i\Bigr).
\label{eq:remainder}
\end{equation}
In particular, $\mathcal L(W_0+tD)-t\,\Linf(D)$ is nonincreasing in $t$ and converges to
$b(D)$ exponentially fast.
\end{theorem}

\begin{proof}
Fix a sample $i$ and abbreviate $s^0=s^0_i$, $q=q_i$, $A=A_i$. Factoring $e^{t\bar q}$ out
of the log-sum-exp gives
\begin{equation*}
\lse(s^0+tq)=t\bar q+\log\sum_{k\in A}e^{s^0_k}
+\log\Bigl(1+\frac{\sum_{k\notin A}e^{s^0_k-t(\bar q-q_k)}}{\sum_{k\in A}e^{s^0_k}}\Bigr).
\end{equation*}
Subtracting $s^0_{y_i}+tq_{y_i}$ and averaging over $i$ gives \eqref{eq:ray} with
$R_D=\frac1n\sum_ir_i$, where $r_i(t)$ is the last logarithm above. Each $r_i$ is a
log-sum-exp of affine functions of $t$ minus a constant, so it is convex, and each exponent
$s^0_k-t(\bar q-q_k)$ with $k\notin A$ decreases in $t$, so $r_i$ is nonincreasing and
non-negative. Since $\bar q-q_k\ge\omega_i$ for $k\notin A$ and $\log(1+x)\le x$, the
upper bound follows. At $t=0$, $r_i(0)=-\log p^0_i(A)$, and $r_i\equiv0$ when $A=[K]$.
\end{proof}

\begin{corollary}[Core radius and floor]\label{cor:core}
Let $W_0$ and $D$ be as in \cref{thm:ray}.
\begin{enumerate}
\item[(a)] If $\Linf(D)>0$, let $r(D)=R_D(0)/\Linf(D)$. Then
\begin{equation}
\mathcal L(W_0+tD)-\mathcal L(W_0)=\Linf(D)\bigl(t-r(D)\bigr)+R_D(t),
\label{eq:core}
\end{equation}
so the loss along the ray lies above the line $\mathcal L(W_0)+\Linf(D)(t-r(D))$ and
approaches it exponentially, and the chord slope equals
$\Linf(D)\bigl(1-r(D)/t\bigr)+R_D(t)/t$.
\item[(b)] If $D\in\mathcal F$, then $y_i\in A_i$ for every $i$, the loss is nonincreasing
along the ray, and it converges to $b(D)$, the cross-entropy of the softmax at $W_0$
restricted to the favoured classes. This limit is zero if and only if $A_i=\{y_i\}$ for
every $i$, that is, if and only if $D$ classifies every sample strictly correctly.
\end{enumerate}
\end{corollary}

\begin{proof}
Evaluating \eqref{eq:ray} at $t=0$ gives $\mathcal L(W_0)=b(D)+R_D(0)$, and subtracting
this from \eqref{eq:ray} gives \eqref{eq:core}. For (b), $\Linf(D)=0$ forces
$\bar q_i=q_{iy_i}$, so $y_i\in A_i$, and each summand of
$b(D)=-\frac1n\sum_i\log\bigl(e^{s^0_{iy_i}}/\sum_{k\in A_i}e^{s^0_{ik}}\bigr)$ is
non-negative and vanishes exactly when $A_i=\{y_i\}$.
\end{proof}

\Cref{cor:core} describes the loss along each ray as a cone rounded at its apex and gives
the size of the rounded region in closed form. The asymptotic wall reaches the height of
the starting point at the distance $r(D)$, which we call the core radius. Its numerator
$R_D(0)$ is the mean surprisal, $-\log p^0_i(A_i)$, of the classes that the direction
eventually favours. At $W_0=0$ with equal biases, $p^0_i(A_i)\ge1/K$, so the core radius is
at most $\log K/\Linf(D)$. At a trained point the softmax is confident, and a random
direction mostly favours classes that receive a small probability, so the core radius is
large. The gap between the asymptotic wall through $W_0$ and the loss is
$R_D(0)-R_D(t)\in[0,R_D(0)]$. Along a floor direction the loss never increases, from any
starting point, and it tends to zero exactly on the directions that separate the samples
strictly, along which gradient descent on separable data moves
\cite{soudry2018implicit,ji2019implicit}. On a two-dimensional slice $W_0+xD_1+yD_2$, these
statements predict a convex slice, chord slopes that increase along every ray towards the
wall slope $\Linf(D_1\cos\vartheta+D_2\sin\vartheta)$, and level sets that become scaled
copies of one convex polygon far from $W_0$.

\section{The backbone}
\label{sec:backbone}

In the layers before the final one, the parameters reach the logits through products of
weight matrices and through rectifications, so the loss is in general neither convex nor
confined to a band around a cone. The question then becomes whether this loss is a lower
envelope of cones, made of several basins that meet along creases, where the slope jumps
by an amount comparable to the slope itself. It therefore reduces to how large the kinks
of the loss of a rectifier network can be along a line in parameter space.

\subsection{Kinks along a line}
\label{sec:kinks}

Let the network $F(x;\theta)$ be composed of affine maps, rectifications, residual
additions and average pooling, where the affine maps include convolutions, fully
connected layers and batch normalization in inference mode with fixed statistics. Let
$\theta\in\R^{n_\theta}$ collect the parameters that move. For
$\theta(t)=\theta_0+t\delta$, write $F_i(t)=F(x_i;\theta(t))$ and
$f(t)=\frac1n\sum_{i=1}^n\ell(F_i(t),y_i)$, with $\ell$ smooth in the logits. The
activation pattern of sample $i$ changes at $t^\ast$ if some rectified unit of that sample
is active on one side of $t^\ast$ and inactive on the other, for all points close to
$t^\ast$, and $S(t^\ast)$ is the set of such samples.

\begin{proposition}[Kinks along a line in parameter space]
\label{prop:kinks}
On every bounded interval, $f$ is continuous and infinitely differentiable except at
finitely many points, at each of which $S(t^\ast)$ is not empty, and
\begin{equation}
f'(t^{\ast+})-f'(t^{\ast-})=\frac1n\sum_{i\in S(t^\ast)}
\nabla_F\ell\bigl(F_i(t^\ast),y_i\bigr)^{\T}
\bigl[\partial_tF_i(t^{\ast+})-\partial_tF_i(t^{\ast-})\bigr].
\label{eq:kink}
\end{equation}
\end{proposition}

\begin{proof}
On an interval on which no earlier unit of a given sample changes its activation, every
pre-activation of that sample is a polynomial in $t$, because the weights are affine in
$t$ and every operation between two rectifications is linear. A polynomial that does not
vanish identically has finitely many roots in a bounded interval, and a rectification
whose input vanishes identically keeps its output at zero. Proceeding through the units in
an order of computation, the interval therefore splits into finitely many pieces, on each
of which every $F_i$ is a polynomial, and $F_i$ is continuous. If the pattern of sample $i$
does not change at $t^\ast$, every unit of that sample computes the same polynomial on both
sides, so its term contributes nothing to the jump. For each remaining term, the chain rule
on either side gives \eqref{eq:kink}.
\end{proof}

\subsection{Kinks along almost every line}
\label{sec:generic}

By \cref{prop:kinks}, a crease would require a comparable fraction of the samples to switch at one point.
Whether samples can switch together depends on algebraic coincidences. For an activation
pattern $\sigma$, which fixes every rectification to the identity or to zero, a unit $a$
and a sample $i$, let $P^\sigma_{i,a}(\theta)$ be the pre-activation of $a$ on $x_i$
computed with the rectifications before $a$ fixed by $\sigma$. It is a polynomial in
$\theta$, since the network with fixed rectifications is affine in the features and
polynomial in the parameters.

\begin{definition}[Shared factors and multiplicity]\label{def:mult}
Let $\Phi$ be the finite set of irreducible factors, up to nonzero constant multiples,
of the polynomials $P^\sigma_{i,a}$ that do not vanish identically. For $\varphi\in\Phi$,
let $\mu(\varphi)$ be the number of samples $i$ for which $\varphi$ divides some nonzero
$P^\sigma_{i,a}$, and let $\mu^\star=\max_{\varphi\in\Phi}\mu(\varphi)$. The data satisfy
the coprimality condition if $\mu^\star=1$.
\end{definition}

\begin{theorem}[Kinks along almost every line]\label{thm:generic}
There is a closed semialgebraic set $Z_2\subset\R^{n_\theta}$ of dimension at most
$n_\theta-2$ such that, if $\theta_0\notin Z_2$, then for almost every direction
$\delta\in\R^{n_\theta}$ we have $|S(t^\ast)|\le\mu^\star$ at every $t^\ast\in\R$. If
$\ell$ is softmax cross-entropy, then
\begin{equation}
\bigl|f'(t^{\ast+})-f'(t^{\ast-})\bigr|\le\frac{\sqrt2\,\mu^\star}{n}
\max_{i\in S(t^\ast)}\bigl\|\partial_tF_i(t^{\ast+})-\partial_tF_i(t^{\ast-})\bigr\| .
\label{eq:jumpbound}
\end{equation}
Under the coprimality condition, every kink along such a line is the share of one sample.
\end{theorem}

\begin{proof}
We use three facts on semialgebraic sets \cite[Sections 2.8 and 2.9]{bochnak1998real}.
First, if $\varphi,\psi\in\R[\theta]$ have no common nonconstant factor, their common real
zero set has dimension at most $n_\theta-2$. The greatest common divisor does not change
when the coefficients are extended to $\mathbb C$, so the complex common zero set has
dimension at most $n_\theta-2$, and its real points have at most this dimension. Second,
the image of a semialgebraic set under a semialgebraic map has at most the dimension of the
set. Third, a semialgebraic subset of $\R^{n_\theta}$ of dimension less than $n_\theta$
has Lebesgue measure zero.

Let $Z_2$ be the union, over all pairs of non-associate $\varphi,\psi\in\Phi$, of their
common real zero sets. It is closed, semialgebraic and of dimension at most $n_\theta-2$.
The directions for which the line meets $Z_2$ at some $t\ne0$ form the image of
$Z_2\times(\R\setminus\{0\})$ under $(v,t)\mapsto(v-\theta_0)/t$, of dimension at most
$n_\theta-1$ and therefore of measure zero. For every other direction, since
$\theta_0\notin Z_2$, the line meets $Z_2$ at no $t\in\R$.

Fix such a direction, a point $t^\ast$ and a sample $i\in S(t^\ast)$. Let $a$ be the first
unit of sample $i$, in the order of computation, whose activation differs between the two
sides of $t^\ast$. The units before $a$ have the same activations on both sides, which
define a pattern $\sigma$, so the pre-activation of $a$ is the same polynomial
$\chi(t)=P^\sigma_{i,a}(\theta(t))$ on both sides. It is positive on one side and not on
the other, so $\chi\not\equiv0$ and $\chi(t^\ast)=0$. Hence some irreducible factor $\varphi_i\in\Phi$
of $P^\sigma_{i,a}$ vanishes at $\theta(t^\ast)$. For two samples $i,j\in S(t^\ast)$, the
factors $\varphi_i$ and $\varphi_j$ vanish at the same point $\theta(t^\ast)\notin Z_2$, so
they are associates. A single $\varphi\in\Phi$ therefore divides a pattern polynomial of
every sample in $S(t^\ast)$, and $|S(t^\ast)|\le\mu(\varphi)\le\mu^\star$. For softmax
cross-entropy, $\nabla_F\ell=p-e_y$ with $\|p-e_y\|^2=(1-p_y)^2+\sum_{k\ne y}p_k^2\le2$, and
the Cauchy--Schwarz inequality in \eqref{eq:kink} gives \eqref{eq:jumpbound}.
\end{proof}

Along almost every line, the training loss averaged over many samples therefore has many
kinks, roughly in proportion to $n$, each carrying the share of at most $\mu^\star$
samples. The coprimality condition is a property of the architecture and the data, and it
fails in two familiar ways. Identical samples have identical pattern polynomials, and two
images that agree on the input patch of a first-layer unit, as images with saturated
regions can, give that unit identical pre-activation polynomials. In both cases a kink
carries the share of the few samples involved. The theorem covers directions drawn from a
distribution with a density. A filter-normalized direction rescales each filter of a
Gaussian direction to a prescribed norm, so its distribution lives on a product of spheres,
a set of measure zero, and the theorem does not apply to it as stated.

\subsection{Creases on sampled values}
\label{sec:sampled}

Individual kinks of this size are too small to be resolved on a grid, so a test on sampled
values must bound what a grid can hide. For values $L_i=L(t_i)$ on a grid $t_i=t_0+ih$, let
$\Delta^2L_i=L_{i+1}-2L_i+L_{i-1}$ and $D(h)=\max_i|\Delta^2L_i|/h^2$. On a smooth function,
$D(h)$ approaches a finite curvature as $h\to0$, whereas across a jump $\Delta s$ of the
slope it grows like $|\Delta s|/h$. If $D(h)$ grows like $h^{-\gamma}$, the exponent
$\gamma$ is therefore close to zero on a smooth function and close to one across a crease.

\begin{lemma}[A bound on hidden creases]\label{lem:floor}
Let $L=S+C$ on the grid, where $C(t)=\Delta s\,(t-t_c)_+$ is a crease at a point
$t_c\in[t_k,t_{k+1}]$ between interior grid points. Then
$\max(|\Delta^2C_k|,|\Delta^2C_{k+1}|)\ge|\Delta s|\,h/2$. If $|\Delta^2S_{i^\ast}|\le
h^2D(h)$ at the index $i^\ast\in\{k,k+1\}$ of this maximum, then $|\Delta s|\le4hD(h)$.
\end{lemma}

\begin{proof}
With $t_c=t_k+\vartheta h$ and $\vartheta\in[0,1]$, we get $\Delta^2C_k=\Delta
s(1-\vartheta)h$ and $\Delta^2C_{k+1}=\Delta s\,\vartheta h$. At $i^\ast$, the triangle
inequality gives $|\Delta s|h/2\le|\Delta^2L_{i^\ast}|+|\Delta^2S_{i^\ast}|\le2h^2D(h)$.
\end{proof}

The hypothesis states that the rest of the function contributes no more at the crease than
the largest second difference on the line. It cannot be checked from the samples alone, so
\cref{sec:num-backbone} calibrates the bound on synthetic lines.

\begin{remark}[Roughness from many small kinks]\label{rem:walk}
Let the slope of $L$ be $s_0+W$, with $s_0$ smooth and $W$ a pure-jump process with jumps
of mean zero and variance $\kappa^2$ at rate $N$ per unit length. Integration by parts gives
$\Delta^2L_i=\int K_h(r-t_i)\,ds(r)$ with the triangular kernel $K_h(r)=(h-|r|)_+$. The
smooth part contributes $O(h^2)$, and the jumps a random term of variance
$N\kappa^2\int K_h^2=\frac23N\kappa^2h^3$. When this term dominates, $D(h)$ grows like
$\kappa\sqrt N\,h^{-1/2}$, up to the slowly varying factor of a maximum over about $1/h$
grid points, so $\gamma$ is close to one half. For kinks that each carry the share of one
sample, $\kappa$ is of order $1/n$ and $N$ of order $n$, so $\kappa\sqrt N$ decreases like
$n^{-1/2}$.
\end{remark}

\section{Numerical illustrations}
\label{sec:numerics}

This section illustrates the statements of \cref{sec:final,sec:backbone} on trained
networks. Section~\ref{sm:examples} gives numerical examples of the theorems on small
instances, and the other sections of the supplement give the details of the measurements.
The final-layer measurements use ResNet-18 \cite{he2016resnet} in its CIFAR form with an
affine head, trained on CIFAR-10 for $100$ epochs by stochastic gradient descent with
momentum $0.9$, a cosine schedule, batches of $128$, weight decay $5\times10^{-4}$ and the
learning rate $0.1$, with batch normalization \cite{ioffe2015bn} and over five seeds. The
backbone measurements use ResNet-56 with and without skip connections
(Section~\ref{sm:protocol}).

\subsection{The final layer}
\label{sec:num-final}

\begin{figure}[tbhp]\centering
\includegraphics[width=0.62\textwidth]{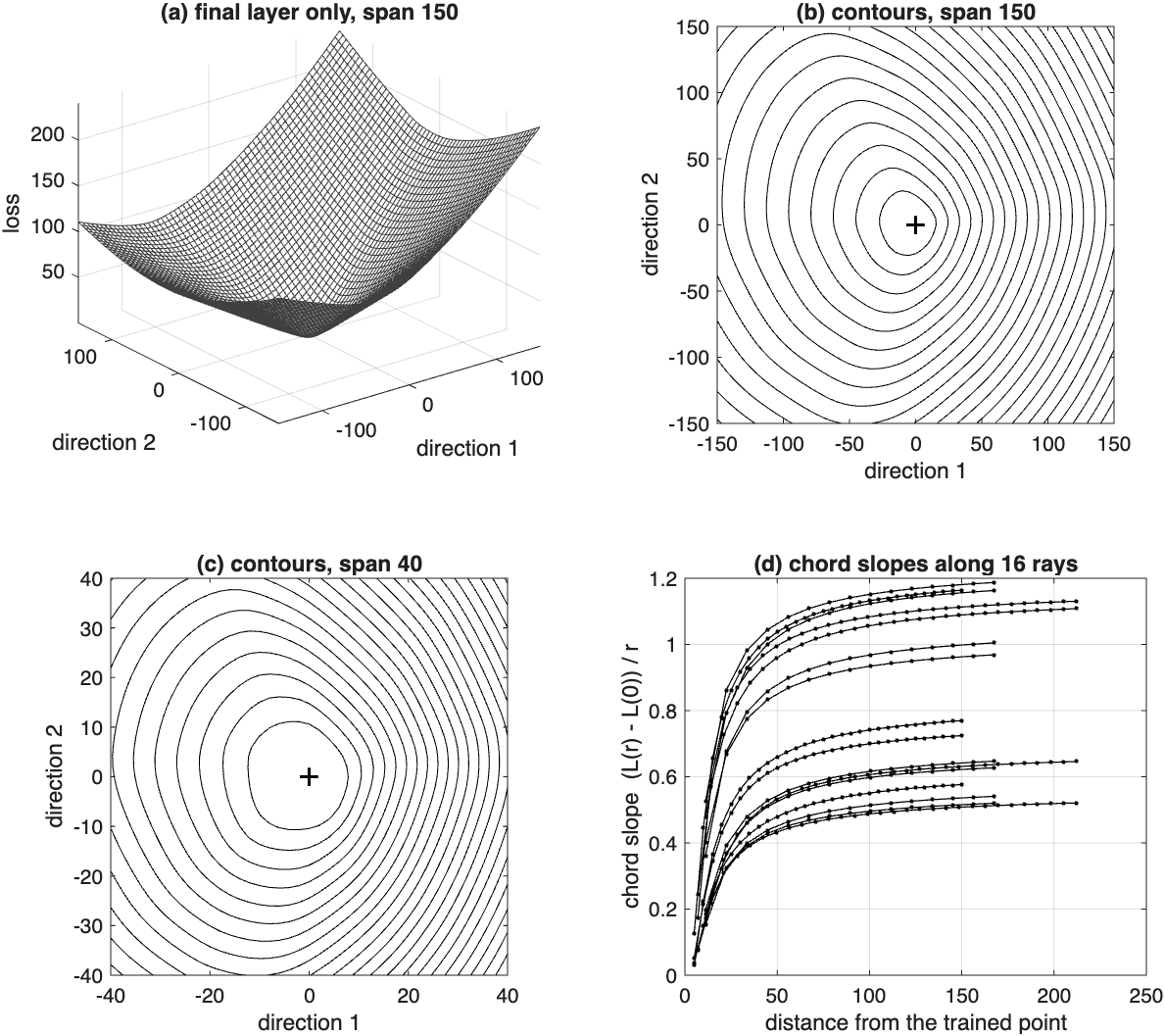}
\caption{The panels show the test loss of ResNet-18 with batch normalization on a plane
through the trained point ($+$), along which only the weights of the affine final layer
vary. Panels (a) and (b) show the surface and its contours over a window of half-width
$150$, and panel (c) shows the contours over a window of half-width $40$, where the
log-sum-exp rounds the apex. Panel (d) shows the chord slope along sixteen rays, which is
nondecreasing and levels off at a wall slope that depends on the direction.}
\label{fig:headslice}
\end{figure}

\Cref{fig:headslice} shows slices of the test loss of ResNet-18 with batch normalization
through the trained weights $W_0$ of its affine final layer, spanned by two random
directions whose rows are rescaled to the norms of the rows of $W_0$, as in the filter
normalization of \cite{li2018landscape}. On both windows the minimum lies at $W_0$, and the
second differences along the axes and diagonals are positive at every grid point, as
convexity requires. Along sixteen rays the chord slope increases strictly and levels off at
values between $0.52$ and $1.19$, and along seventy-two rays the exponent $p$ for which
$\mathcal L-\mathcal L(W_0)$ grows like $t^p$ has median values $3.07$, $1.97$, $1.43$,
$1.20$ and $1.10$ in the bands $[5,10]$, $[10,20]$, $[20,40]$, $[40,80]$ and $[80,150]$ of
the distance $t$, approaching the value one of a straight wall. The level set at height
one of the asymptotic cone, reconstructed from the limiting chord slopes, is convex.

A rounded cone $\sigma\bigl((t^2+r_c^2)^{1/2}-r_c\bigr)$ fitted along each of the
seventy-two rays approaches $\sigma(t-r_c)$ for large $t$, so \cref{cor:core} identifies
$\sigma$ with $\Linf(D)$ and $r_c$ with $r(D)$. On the slice of \cref{fig:headslice}, the
fitted wall slopes lie between $0.55$ and $1.26$, matching the limiting chord slopes of
$0.54$ to $1.25$, and the fitted core radii lie between $8.3$ and $14.4$, with median
$12.0$. To compare such fits with exact values, we evaluate the expansion of
\cref{thm:ray} in closed form on the same checkpoint, the same $1024$ test images and the
same two directions. Along all seventy-two rays, the remainder $R_D$ stays between zero and
the bound \eqref{eq:remainder} at $600$ distances up to $150$. The wall slopes $\Linf(D)$
lie between $0.546$ and $1.250$, and a rounded cone fitted to the exact loss recovers them
to within $2\%$ and reproduces the fitted core radii, between $8.26$ and $14.45$ with median
$11.95$. The mean surprisal $R_D(0)$ lies between $6.51$ and $10.02$, which corresponds to
geometric mean probabilities between $4\times10^{-5}$ and $1.5\times10^{-3}$ on the
favoured classes. It is smallest along the directions that favour the true class on many
images, up to $31\%$ of them, because the trained softmax gives the true class a
probability close to one. The same directions have the shallowest walls, because such
images add nothing to $\Linf(D)$, so the surprisal and the wall slope rise together, with
a correlation of $0.93$ over the rays. The exact core radius $r(D)$ therefore varies less
than the wall slope, between $7.35$ and $13.00$ with median $10.47$. The fitted radius
exceeds it on $70$ of the $72$ rays, by a median of $13\%$ and at most $24\%$, as the
synthetic layers of Section~\ref{sm:examples} lead one to expect, so a fitted core radius
is a biased estimate of the exact one. The loss stays below the asymptotic wall through
$W_0$ by at most $10.00$, within the bound $R_D(0)$ of \cref{cor:core}. A repeat of this
run on a second machine gives exact core radii between $9.04$ and $16.93$, which its
fitted radii exceed by $9$ to $28$ percent.

The floor of the cone is narrow. On the training features, $\Linf$ vanishes along the
direction of $W_0$ itself in four of the five trained networks and stays below
$6\times10^{-4}$ in the others, but none of the $72$ rays of the plane and none of $200$
further filter-normalized directions lies in the floor, so a slice through $W_0$ meets the
walls in every direction. The band of \cref{prop:finallayer} has width between $2.37$ and
$2.66$ at the origin and between $14.0$ and $55.7$ at the trained weights, so the closest
uniform description of the final-layer loss is a rounded cone with its apex at the origin.

\subsection{The backbone}
\label{sec:num-backbone}

A small rectifier network with two hidden layers of $64$ units and ten classes, in which
every breakpoint along a filter-normalized line is the root of a known polynomial,
illustrates \cref{prop:kinks,thm:generic}. As $n$ grows from $16$ to $128$ and $1024$, the
number of kinks per unit length grows from $390$ to $3114$ and $24\,719$, while the median
kink, relative to the mean slope, falls from $2.4\times10^{-3}$ to $2.8\times10^{-4}$ and
$3.4\times10^{-5}$, roughly in proportion to $1/n$. One-sided derivatives at well-isolated
kinks reproduce \eqref{eq:kink} to relative errors between $6\times10^{-6}$ and
$4\times10^{-3}$. Distinct samples never share a breakpoint, whereas a duplicated sample
shares all of its breakpoints with its copy.

\begin{figure}[tbhp]\centering
\includegraphics[width=0.72\textwidth]{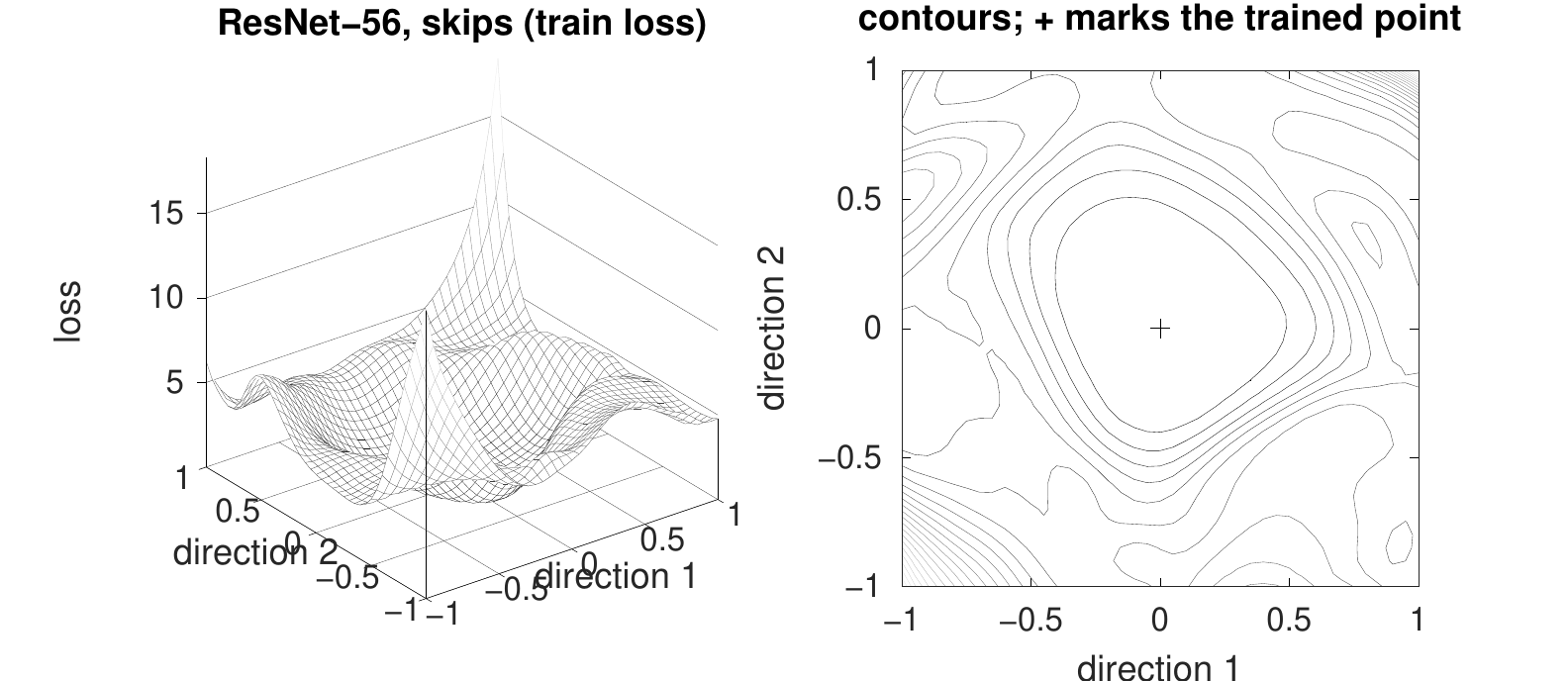}\\[2pt]
\includegraphics[width=0.72\textwidth]{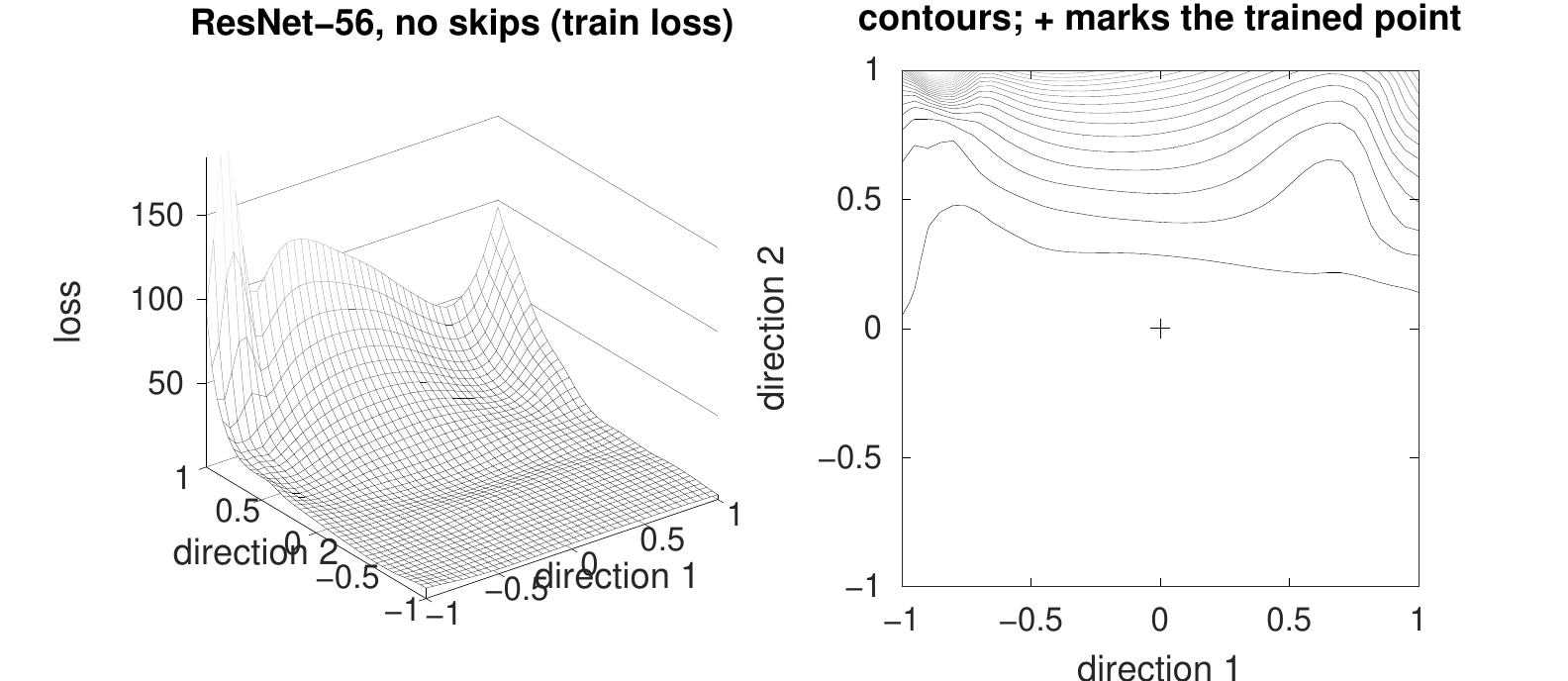}
\caption{The panels show training-loss planes through the whole backbone of ResNet-56
with skip connections (above) and without them (below), along two
filter-normalized directions with batch normalization in inference mode. Each slice has a
single basin, and the plain network is considerably steeper and lopsided. The wrinkles on
the walls contain no creases, as \cref{tab:crease} shows.}
\label{fig:r56}
\end{figure}

\Cref{fig:r56} shows planes of the training loss through the whole backbone of ResNet-56
with and without skip connections, along two filter-normalized directions, on $512$
training images with batch normalization in inference mode. In each slice, the trained
point is the minimum of a single basin, and the same holds in twenty further planes
(Section~\ref{sm:crease}), including those of the plain network, whose loss surface Li
\emph{et al.} describe as chaotic for their networks \cite{li2018landscape}. The walls
carry small wrinkles. Increasing the number of evaluation images eightfold leaves their
amplitude unchanged, at a ratio of $0.98$, and their positions fixed, at a correlation of
$0.994$, whereas by \eqref{eq:kink} the kinks become about eight times smaller, so the
wrinkles are undulations of the function computed on the images as a whole. In six refined
windows, the exponents $\gamma$ of \cref{tab:crease} stay far from the value of an envelope
of cones. On forty lines with $257$ points over a half-width of $0.02$, the activation
patterns change about $10^8$ times per unit length, and $\gamma$ has median $0.50$, with a
bootstrap $95\%$ interval from $0.44$ to $0.60$, as \cref{rem:walk} predicts for dense small
kinks. On synthetic lines, the exponents of such random walks and of single creases
overlap between about $0.6$ and $0.9$, so we bound the size of any crease directly with
\cref{lem:floor}, whose bound held on all $4000$ synthetic lines with a crease of known
size. On the walls, the bound is at most
$2.5\%$ of the local slope with skip connections, and without them it is $1.6\%$ on the
median line and $29\%$ on the least favourable one. A crease of an envelope of cones, which
changes the slope by an amount comparable to the slope itself, is therefore absent from
every line, and the correspondence with cones ends at the backbone.

\begin{table}[tbhp]\centering
\caption{Scaling exponent $\gamma$ of the normalized second difference $D(h)$ in six
refined windows, with two control surfaces.}\label{tab:crease}
\small
\begin{tabular}{@{}lccc@{}}
\toprule
 & dimple & plain wall & basin floor\\
\midrule
ResNet-56, with skips & 0.29 & 0.56 & 0.55\\
ResNet-56, no skips   & 0.40 & 0.27 & 0.15\\
\midrule
\multicolumn{4}{@{}l}{controls: smooth ripple $0.14$ \quad lower envelope of cones $1.16$}\\
\bottomrule
\end{tabular}
\end{table}

\section{Discussion and conclusion}
\label{sec:discussion}

At the final layer, softmax cross-entropy is a rounded polyhedral cone in an exact sense.
Along every ray from any starting point, the loss equals a linear term, whose slope is the
recession function, plus a constant in closed form and a remainder that decreases
monotonically and exponentially. The core radius expresses the size of the rounded region
through the confidence of the softmax at the starting point. This explains why the band of
\cref{prop:finallayer} is narrow at the origin and wide at a trained point. The exact
expansion also shows that a rounded cone fitted to sampled values overestimates the core
radius.

In the backbone, the conic picture ends. Along almost every line, each kink of the loss
carries the share of at most $\mu^\star$ samples, so the kinks become smaller in proportion
to $1/n$ while their number grows in proportion to $n$, and the creases of an envelope of
cones do not form. The wrinkles on the slices of ResNet-56 keep their size and position
when the number of images grows, so they are undulations of the function that the network
computes on the images as a whole. On the sampled lines, the bound of \cref{lem:floor}
limits any hidden crease to at most $2.5\%$ of the local slope with skip connections, and
to $1.6\%$ on the median line and $29\%$ on the least favourable line without them.

\Cref{thm:ray,thm:generic} are exact, and the measurements that illustrate them use slices
along filter-normalized directions of ResNet-18 and ResNet-56 on CIFAR-10. Two extensions
are left for further study. \Cref{thm:generic} covers directions drawn from a distribution
with a density, and its extension to filter-normalized directions, which lie on a product
of spheres, is open. The results for the final layer require logits that depend linearly on
the weights, and a head with nonlinear scores, such as the conic head of the companion
paper \cite{gerek2026margins}, is not covered by them.

\section*{Code availability}
The code that trains the networks and reproduces every number and figure is available at
\url{https://github.com/ongerek/Conic}.

\section*{Use of generative AI}
Generative AI tools from Anthropic's Claude models were used for the development of the
code, the verification of the mathematical formulations, the typesetting of mathematical
formulations, and grammatical corrections and paraphrasing.

\section*{Appendices}

\appendix

These supplementary materials contain numerical examples of the statements of the paper,
the training protocol, and the measurements on the backbone that the paper summarizes.

\section{Numerical examples of the theorems}
\label{sm:examples}

Following the proofs of Theorems~\ref{thm:ray} and~\ref{thm:generic} and their corollaries, we give
numerical examples below. They illustrate each result on small instances in which every
quantity can be computed to high accuracy.

\paragraph{The final layer} In $200$ random instances with $n$ between $5$ and $59$, $d$
between $2$ and $11$ and $K$ between $2$ and $7$, a quarter of them with directions that
produce ties among the favoured classes, the remainder $R_D(t)$ of (\ref{eq:ray}) lies
between zero and the bound of (\ref{eq:remainder}) at $111$ values of $t$ between $0$ and
$500$, and $\mathcal L(W_0+tD)-t\Linf(D)$ is nonincreasing in every instance. To see how
the core radius of Corollary~\ref{cor:core} relates to a fitted rounded cone, we train an affine
layer by gradient descent on rectified Gaussian features of dimension $512$ with ten
classes, evaluate the loss on $1024$ held-out features along filter-normalized directions,
and fit $\sigma\bigl((t^2+r_c^2)^{1/2}-r_c\bigr)$ over $t\in[0,150]$. We vary the
separation of the classes and the length of training over six settings, which give
training losses at $W_0$ between $0.001$ and $0.06$ and exact core radii $r(D)$ between
$2.5$ and $10.9$. Over $128$ directions, the fitted slope $\sigma$ agrees with $\Linf(D)$
to within $1.2\%$, and the fitted radius $r_c$ exceeds $r(D)$ by $3.5$ to $17$ percent,
with medians that grow from $5\%$ in the settings with the smallest radii to $13\%$ in the
setting with the largest. The fitted radius is thus an overestimate of $r(D)$, because the
rounded cone and the expansion (\ref{eq:core}) round the apex in different ways.

\paragraph{Breakpoints} In the small rectifier network of Section~\ref{sec:num-backbone}, with
$64$ samples drawn from ten Gaussian clusters and the breakpoints of every sample computed
exactly as roots of polynomials on $t\in(0.5,1.5)$, the $1574$ breakpoints of distinct
samples are distinct, the smallest distance between breakpoints of two samples being
$6.1\times10^{-8}$. When one sample is duplicated, the two copies share every breakpoint,
as Definition~\ref{def:mult} anticipates.

\section{Training protocol}
\label{sm:protocol}

The final-layer measurements use ResNet-18 \cite{he2016resnet} in its CIFAR form with an
affine head. Training on CIFAR-10 \cite{krizhevsky2009cifar} runs for $100$ epochs with
stochastic gradient descent, momentum $0.9$, a cosine schedule, batches of $128$, weight
decay $5\times10^{-4}$ and the learning rate $0.1$, with batch normalization
\cite{ioffe2015bn}, over five seeds. The slices of Section~\ref{sec:num-final} pass through the
trained weights $W_0$ of the affine final layer and hold the backbone, and hence the
features, fixed.

The backbone measurements use ResNet-56 \cite{he2016resnet} in its CIFAR form, with and
without its skip connections, trained on CIFAR-10 under the same recipe. A random
direction in parameter space is filter-normalized as in \cite{li2018landscape}: each
filter of a Gaussian direction is rescaled to the norm of the corresponding filter of the
trained weights, so that the apparent scale of the slice does not depend on the scale of
the weights. A slice through a trained point $\theta_0$ along two such directions
$\delta_1,\delta_2$ evaluates the training loss on the grid
$\theta_0+x\delta_1+y\delta_2$, with batch normalization in inference mode and fixed
statistics, as Proposition~\ref{prop:kinks} assumes. The loss is averaged over $512$ training images,
read without augmentation, so that every grid point sees the same images. A repeat of the
final-layer run on a second machine, a workstation with an NVIDIA RTX PRO 6000 Blackwell
GPU, reproduces the measurements of Section~\ref{sec:num-final}.

\section{Backbone planes and refined lines}
\label{sm:crease}

We repeated the test of Section~\ref{sec:num-backbone} on ten further planes for each network,
drawn with filter-normalized directions over all parameters, the final layer included. In
each of the twenty planes, the trained point is the minimum of a $21\times21$ slice over
the same window, and the slice contains a single deep basin. At radius $0.9$, the loss
ranges from $2.9$ to $12.0$ with skip connections and from $3.6$ to $47.1$ without them.
In the published slice of the plain network, the loss at radius $0.9$ ranges from $3.1$
to $68.5$ depending on the direction.

In each plane, we refined two lines along the first direction, one through the trained
point and one across a wall at radius $0.5$. Each line has $257$ points over a half-width
of $0.02$, and the loss is computed in double precision. Counting the changes of the
activation pattern between neighbouring points gives about $10^8$ kinks per unit length,
which is about $1.6\times10^4$ kinks between two neighbouring points of the grid. Over the
$40$ lines, the exponent $\gamma$ has median $0.50$, with a bootstrap $95\%$ interval from
$0.44$ to $0.60$, and it ranges from $0.19$ to $0.87$.

We evaluated synthetic lines, made of a parabola and a random walk of the slope, on the
same grid and with the same estimator. As the weight of the random walk grows from zero,
the median exponent rises from $0$ through about $0.2$, $0.4$ and $0.5$ to $0.6$, and
single lines spread from about $0.1$ to $0.9$. A single crease, placed at a random
position between grid points, gives a median near $0.9$ and a spread from about $0.6$ to
$1.2$. In $4000$ synthetic lines with a crease of known size, the bound $4hD(h)$ of
Lemma~\ref{lem:floor} held in every case. On the walls of the network planes, the bound is at
most $2.5\%$ of the local slope with skip connections. Without them, it is $1.6\%$ of the
local slope on the median line and $29\%$ on the least favourable one. On the basin floor,
where the mean slope vanishes and a comparison with it has no meaning, the bound is at
most $2.2\times10^{-4}$ times the slope of the wall in the same plane with skip
connections and at most $0.084$ times it without them.

\bibliographystyle{abbrv}
\bibliography{references,companion_refs}

\end{document}